# Enhanced Direct and Indirect Genetic Algorithm Approaches for a Mall Layout and Tenant Selection Problem

Uwe Aickelin
School of Computer Science
University of Nottingham
NG8 1BB   UK
uxa@cs.nott.ac.uk

Kathryn A. Dowsland, *European Business Management School, University of Wales Swansea, Singleton Park, Swansea SA2 8PP, UK.*

Uwe Aickelin and Kathryn A Dowsland are both members of the Operational Research Society.

Correspondence to Uwe Aickelin.

**Abstract**

During our earlier research, it was recognised that in order to be successful with an indirect genetic algorithm approach using a decoder, the decoder has to strike a balance between being an optimiser in its own right and finding feasible solutions. Previously this balance was achieved manually. Here we extend this by presenting an automated approach where the genetic algorithm itself, simultaneously to solving the problem, sets weights to balance the components out. Subsequently we were able to solve a complex and non-linear scheduling problem better than with a standard direct genetic algorithm implementation.

This paper describes a genetic algorithm approach to a mall layout and tenant selection problem, in future mall problem for short. Two types of genetic algorithms are presented. Firstly, a direct approach where the encoding of the strings represents a one to one relationship with the layout of the mall. The second algorithm is an indirect approach where the genetic algorithm finds good permutations of the locations, which are then filled by a separate decoder routine. The work presented here is an extension of our earlier research described in Aickelin and Dowsland (1999 and 2000). We will confirm here the conclusions that the indirect approach is more powerful than its direct counterpart even for the more complex and non-linear mall problem. Furthermore, we will investigate further enhancements of the indirect approach to find a balance between striving for feasibility and good target function values within the decoder.

The mall problem arises both in the planning phase of a new shopping centre and on completion when the type and number of shops occupying the mall has to be decided. To maximise revenue a good mixture of shops that is both heterogeneous and homogeneous has to be achieved. Due to the difficulty of obtaining real-life data because of confidentiality, the problem and data used in this research are constructed artificially, but closely modelled after the actual real-life problem as described for instance in Bean et al. (1988). In the following, we will briefly outline our model.

The objective of the mall problem is to maximise the rent revenue of the mall. Although there is a small fixed rent per shop, a large part of a shop's rent depends on the sales revenue generated by it. Therefore, it is important to select the right number, size and type of tenants and to place them into the right locations to maximise revenue. As outlined in Bean et al. (1988), the rent of a shop will depend on the following factors:

- The attractiveness of the area in which the shop is located.
- The total number of shops of the same type in the mall.
- The size of the shop.
- Possible synergy effects with neighbouring shops of the same group (not used by Bean et al.).
- A fixed amount based on the type of the shop and the area in which it is located.

Before placing shops in the mall, the mall is divided into a discrete number of locations, each big enough to hold the smallest shop size. Larger sizes can be created by placing a shop of the same type in adjacent locations. Hence the problem is that of placing $i$ shop types (e.g. menswear) into $j$ locations, where each shop type can belong to one or more of $l$ groups (e.g. clothes shops) and each location is situated in one of $k$ areas. For each type of shop there will be a minimum, ideal and maximum number allowed in the mall, as consumers are drawn to a mall by a balance of variety and homogeneity of shops.

The size of shops is determined by how many locations they occupy within the same area. For the purpose of this study, shops are grouped into three size classes, namely small, medium, and large, occupying one, two, and three locations in one area of the mall respectively. For instance, if there are two locations to be filled with the same shop type within one area, then this will be a shop of medium size. If there are five locations with the same shop type assigned in the same area, then they will form one large and one medium shop etc. Usually, there will be a maximum total number of small, medium and large shops allowed in the mall.

To test the robustness and performance of our algorithms thoroughly on this problem, 50 problem instances were created. The data was grouped into five sets (numbered 3-7) of increasing difficulty with ten problem instances each. All problem instances have 100 locations grouped into five areas. However, the sets differ in the number of shop types available (between 50 and 20) and in the tightness of the constraints regarding the minimum and maximum number of shops of a certain type or size. Furthermore, for a better comparison of results, the problem instances in sets 4 to 7 are set up identical apart from the tightness of the constraints. Full details on how the data was created, its dimensions, the differences between the sets and the anticipated effects on our algorithms can be found in Aickelin (1999).

**1. Genetic Algorithms and Set-up of Experiments**

Genetic algorithms are generally attributed to John Holland (1976) and his students in the 1970s, although evolutionary computation dates back further (refer to Fogel (1998) for an extensive review of early approaches). Genetic algorithms are stochastic meta-heuristics that mimic some features of natural evolution. Canonical genetic algorithms were not intended for function optimisation, as discussed by De Jong (1993). However, slightly modified versions proved very successful. For an introduction to genetic algorithms for function optimisation, see Deb (1996).

In short, genetic algorithms mimic the evolutionary process and the idea of the survival of the fittest. Starting with a population of randomly created solutions, better ones are more likely to be chosen for recombination into new solutions, i.e. the fitter a solution, the more likely it is to pass on its information to future generations of solutions. In addition to recombining solutions, new solutions may be formed through mutating or randomly changing old solutions. Some of the best solutions of each generation are kept whilst the others are replaced by the newly formed solutions. The process is repeated until stopping criteria are met.

To obtain statistically sound results all experiments were conducted as twenty runs over all 50 problem instances. All experiments were started with the same set of random seeds, i.e. with the same initial populations. The platform for experiments was a Pentium 200 MMX based IBM compatible PC, run in DOS 7.0. All algorithms are coded in Turbo Pascal for Dos 7.01. The results are presented in feasibility and rent format. Feasibility denotes the probability of finding a feasible solution averaged over all problem instances. Rent refers to the objective function value of the best feasible solution for each problem instance averaged over the number of instances for which at least one feasible solution was found. Should the algorithm fail to find a single feasible solution for all 20 runs on one problem instance, a censored observation of zero is made instead. As we are maximising the rent, this is equivalent to a very poor solution. The values for the rent in the figures are in thousands of pounds.

As the mall problem is non-linear and hence unsuitable for traditional methods, the optimal solutions are unknown. However, to get an estimate as to the quality of solutions presented here, the following theoretical upper bound can be used. The best one can hope for is that all shops are large, in a group and with an ideal size count. This is obviously too optimistic, as some problem instances do not allow large shops only and many of the other constraints will prevent an ideal shop count for all shop types. For instance, there is no rule that the sum of the ideal shop counts must be equal to the total number of locations. In the ideal case, the rent for a mall with 100 locations would be 2640. Full details how this bound was calculated can be found in Aickelin (1999).

## 2. Direct Genetic Algorithm Approach

As a benchmark for future experiments, a straightforward direct genetic algorithm approach to the mall problem is presented in this section. In order to do this, a suitable encoding must be determined. The following two possibilities were considered. In the first, the string has as many genes as the number of shop types multiplied by the number of areas. Each gene then denotes the number of shops of a particular type built in a specific area. The second possibility is a string with as many genes as the number of locations in the mall. Here, the value of a gene would indicate the type of shop built in a specific location.

With the first encoding, there are a number of problems, which led us to discard the idea. The biggest problem is that after any type of standard crossover, we will almost certainly end up with infeasible children in respect to the total number of shops used. Thus, an intensive repair operator would be required. Furthermore, for some data files we would end up with very long strings of up to 250 genes. This could lead to problems regarding the formation of successful building blocks. An advantage of this encoding would be the quicker calculation of the objective function value, as the $n_{jk}$ values (number of shops of one type per area), which determine the sizes of the shops, are already known. With the second encoding, the $n_{jk}$ values have to be deducted first.

The second encoding offers advantages over the other type for the mall problem. First, crossover operations will never generate infeasible children regarding the total number of shops. Solutions can still be infeasible with respect to the shop size and shop count constraints, but this is the same with the first encoding. Secondly, strings are no longer than 100 genes for any of the data files. Thus, for its general superiority and simplicity the second encoding was chosen.

Since only the multiple-choice constraints are implicitly taken care of by the chosen encoding, we have to deal with the remaining constraints either with a repair algorithm or a penalty function. These constraints are the minimum and maximum number of shops of each type and the maximum number of small, medium and large sized shops allowed in the mall. We decided against using a repair mechanism, as no intuitively obvious algorithm is available. Thus finding a suitable problem specific routine would be a new optimisation problem, nearly as difficult as solving the original problem. Instead, a penalty function approach was implemented. To arrive at the penalty $p_s$ of a solution $s$, we will measure and sum the violations of all these constraints and then multiply this with a penalty weight $w_{penalty}$. If a constraint is violated, the violation is measured as the difference between the actual number and the minimum respectively maximum number allowed. The raw fitness of an individual can then be calculated as its objective function value minus the above penalty term.

With this in place, a standard canonical genetic algorithm as described for instance in Aickelin and Dowsland (1999) can be used in the first instance. Following initial experiments reported in Aickelin (1999), the genetic algorithm uses the parameters and strategies as summarised in Table 1. Using these parameters and genetic operators, the genetic algorithm was able to solve the mall problem reasonably well. The results are displayed in Figure 1 under the 'direct' label. Although the optimal solutions are unknown, an average feasibility of above 90% seems to indicate that we have found solutions of at least reasonable quality. Furthermore, the solutions found were within 30% of a very optimistic upper bound.

## 3. Indirect Genetic Algorithm Approach

In this section, we will describe a different type of approach for the mall problem, which had been used successfully for a nurse scheduling problem in Aickelin and Dowsland (2000). There it was concluded that to be successful the decoder has to strike a balance between being an optimiser and striving for feasible solutions. In contrast to the direct genetic algorithm solving the actual layout problem, the 'indirect' genetic algorithm finds an optimal permutation of locations, which is then fed into a decoder. This decoder builds the actual solution, in this case the layout of the mall, from the given permutations. A permutation of locations was chosen for the encoding used for the indirect genetic algorithm approach. This decision was made because this type of encoding is more compact and allows for a simpler decoder than the other alternatives available.

Having decided on an encoding, the remainder of the indirect genetic algorithm can be set up. This will be along the same lines as for the direct genetic algorithm in the previous section and the indirect approach for the nurse scheduling in Aickelin and Dowsland (2000). In other words, we use those strategies and parameters that in general proved best for the indirect genetic algorithm, which are parameterised uniform permutation crossover (PUX) with *p=0.66* and standard swap mutation. PUX works in a similar way to uniform order based crossover as described in Syswerda (1996). However, when creating the binary template, the probability for a one will be equal to a parameter *p*, similar to standard parameterised uniform crossover. All problem specific parameters and strategies used, such as the penalty weight and penalty function, are those that worked best in the direct approach. As the decoder cuts out large unattractive parts of the solution space, a significant reduction in population size was possible. This cancelled out the otherwise large increase in computation time due to the use of the decoder itself. Some brief empirical tests confirmed this intuitive decision, showing that solution quality did not further improve for population sizes larger than 100. A summary of all parameters is shown in Table 1.

As outlined earlier, the task of the decoder is to assign shop types to locations in the order determined by the genetic algorithm. To decide which shop is best placed into the location currently under consideration, the decoder will cycle through all possible shop types. Those shop types for which the maximum number of shops allowed in the mall are already present will not be considered. Each candidate shop type is temporarily placed into the location and the following points are then taken into account to calculate its suitability:

- What is the fixed rent for this shop type and location?
- How many shops of this type are already present in the mall and how does this compare to the minimum, ideal and maximum number allowed for this shop type?
- Would placing this shop (help) complete a group or is its group already complete?
- What shop size would be created by placing the shop and how would this affect the total number of small, medium and large shops allowed?

The decoder then assigns a score for each candidate according to its performance on the above points. The candidate with the highest score, or in the case of a tie, the first one with the highest score is placed into the location. Note that ties are extremely unlikely due to the partial randomness of the data. Formally, the score $s_{ij}$ of a shop of type $i$ going into location $j$ is calculated as follows, where $w_1$-$w_6$ are appropriate weights:

$$s_{ij} = [w_1 B_m + w_2 B_l + w_3 S + w_4 I + w_5 M + w_6 G + \text{fixed rent}]$$

Where the meaning of the terms in brackets is as follows:

- The 'medium bonus' $B_m$ and 'large bonus' $B_l$ are set to 1 if the shop would create the respective shop size, otherwise they are 0.

- The 'slackness in size constraint' $S$ is measured as the number of shops of the size, which would be created, that are still allowed in the mall. For instance, if a small shop was created and there is a total of 5 small shops allowed in the mall, with 3 small shops already present, then $S = 5 - 3 - 1 = 1$. Note that this can lead to negative values for unsuitable shop sizes.

- The 'ideal / total number of shop type' $I$ is equal to the difference between the ideal and the total number for shops of this type if a shop of this type was placed. This encourages shops of those types to be placed, which are still below their ideal level, with the further below they are the higher the encouragement. Note that this should force the shop count for each type to be above the minimum and help it to remain below the maximum (due to minimum ≤ ideal ≤ maximum).

- The 'new member' $M$ is 0 if a shop of this type is already present in the area. Otherwise, it is calculated as $M = (10 - \text{total members of group the shop type is in} + \text{members already present in area})$. This encourages shops of those types whose group is more complete than others to be placed. Ten was chosen, as this is the largest possible group size. Thus, smaller groups are at an advantage, which is intentional as they are more likely to be able to reach completion than larger groups. If a shop type belongs to more than one group, two scores are possible. If a shop type is in no group then $M = 0$.

- The 'group complete' $G$ is set to 1 if the group the shop type belongs to is complete in the area; otherwise it is set to 0. For shop types that belong into more groups, $G$ is set equal to the total number of complete groups the shop type belongs to. If the shop type is in no group then $G = 0$.

- The 'fixed rent' is equal to fixed amount of rent depending on the shop type $i$ and area $k$ the location is in.

Although we have created a powerful scoring mechanism taking all the problem details into account, there is one dilemma: How are the weights going to be set so that there is a good balance not only between the various target function components but also with regards to finding feasible solutions? In a first attempt, three sets of weights (low, medium, high) as detailed in Table 2, were tried. These were based on intuition and previous experience. Note that the labels refer to the relative 'magnitude' of the weights.

An example of a string with ten locations using the low weight settings might therefore look as follows:

(3, 5, 6, 1, 10, 9, 2, 4, 8, 7; 500, 1000, 100, 200, 200, 2000)

The string lists first the permutation of locations followed by the weights used. Thus, the decoder would first try to find the most suitable shop type for location 3. To evaluate which shop type is best for this particular location, the score, as outlined above, for each possible shop type would be calculated. In doing so, the decoder makes use of the weights given in the second part of the string. Amongst all possible shop types, the one with the highest score would be chosen. Next, the decoder would repeat this for location 5. Again, it would cycle through all possible shop types eventually choosing the one scoring highest with the weights provided. This time, the decision will be based on the shop type for location 3 already being fixed. The remaining locations would then be filled in similar fashion in the order given, always basing the decision on the weights provided and the fact that preceding locations in the string are fixed to certain shop types.

The results for these three weight sets can be seen in Figure 1 along with a comparison of the direct genetic algorithm results as described earlier. For the 'low' setting they are about as good as the direct genetic algorithm results, for the 'medium' setting, they are better and for the 'high' setting, they are far worse. Clearly, there is potential in the decoder, if only one could find the 'best' or 'most balanced' weights. However, unlike for the nurse scheduling decoder weights, there are far too many possibilities for these decoder weights to conduct empirical tests. Thus, we will introduce an 'automatic' approach in the next section where the genetic algorithm sets the weights simultaneously whilst optimising the problem.

## 4. Further Decoder Enhancements

The ability of genetic algorithms to set their own penalty parameters has been investigated by many researchers leading to improved results over fixed penalty weights. Examples of this can be found in Reeves (1997), Hadj-Alouane and Bean (1997) and Smith and Tate (1993). Because of the high number of weights, we have to take a slightly different approach here. When initialised, each individual has a random set of weights $w_1...w_6$ assigned as six extra genes. Crossover and mutation is applied to these six extra genes so that weights producing better results have a better chance of being passed on. After the success of the self adjusting decoder weights, we go one step further and self adjusting crossover strategies and mutation rates are introduced. Such genetic algorithms that set their own parameters 'online' are not a new concept. Early approaches (1960s and 1970s) to parameter adaptation are found in the Evolution Strategy area of research as summarised by Fogel (1998). In the evolutionary algorithm field, Tuson and Ross (1998) experiment with co-evolving operator settings in genetic algorithms. Further examples of adapting parameter ideas are presented in Davis (1985) and Yeralan and Lin (1994).

To execute our idea, we attached six genes to each individual, which represent the six decoder weights for this particular individual. However, these extra genes are not treated as part of the string but have their own crossover and mutation operator. The values for the weights were randomly initialised in the range between 0 and 10000. This large range was chosen, as we did not know in which region good values were lying. PUX was applied to the permutation part of the string as before. For the weights part of the string we tried the following three crossover strategies for the weights of the children:

- Taking the weights of a random parent.
- Taking the weights as rank-weighted averages of both parents, i.e. the weight that is passed on will be closer to that of the relatively fitter parent, with the larger the difference in fitness respectively rank between the two parents the closer the weight to the fitter individual's weight. Rank-weighted averages rather than direct fitness-weighted averages are used to avoid scaling issues.
- Setting the weights at random in the range between the two weights of the parents.

Experiments showed that the last two methods performed well with the second method being slightly better and converging more quickly. The results for the second method are shown in Figure 1 under the 'Auto' label. They were the best found so far with 100% feasibility and improved rent. The first method did not produce as good results as the others, although still better than those found by the direct algorithm. We assume this was because this method was restricted to keeping to the weight values it was initialised at, which makes it less flexible than the other two methods.

We also wanted to make sure that the choice of 10000 as the upper initialisation limit did not have a negative effect on the optimisation. This could occur because the fixed rent component of the score has a fixed weight of one assigned to it. Therefore, the algorithm was rerun with an initialisation limit of 50000 and a different random number stream to avoid mere scaling. The results were of the same quality as before. When we took a closer look at the average weights of the final generations, we discovered some interesting similarities. In Figure 2, these weights are pictured as summarised by data sets.

As can be seen from the graphs, the weights behave in a very similar fashion for both choices of initialisation ranges. This indicates that the fixed rent has little to no effect on the decoder assignments. Furthermore, in 'easier' data sets, relatively higher weights were put on rent-increasing rules rather than on rules concerned with constraints. For data sets that are more difficult, the picture is different. Now more emphasis, or a relatively higher weight, is put on the slackness and ideal rules, which both deal with the constraints. These results are very encouraging, as they show that the decoder behaves as intended, striking a good balance between feasibility and quality of results.

The adaptive crossover and mutation idea follows a similar pattern as before. Each individual receives two additional genes. The first is set to 1, 2 or 3 which indicates that respectively a C1 (Reeves (1996)), PMX (Goldberg and Lingle (1985)) or PUX with *p=0.66* crossover is performed. The second gene is a real number giving the swap mutation rate applied to the individual. The crossover tag is initialised at random, such that there is an equal probability for each crossover operator. The swap mutation rate is set randomly between 0% and 5%. When a crossover is performed, the child takes its crossover tag from the parent with the higher rank and the mutation probability of the child is set to the rank-weighted averages of the parents.

Figure 1 shows the results for adaptive crossover alone (label 'Cross') and for adaptive crossover and mutation (label 'Mutat'). As can be seen, the use of adaptive crossover improves results, whilst the addition of adaptive mutation makes them worse. Whilst we will investigate the former in the remainder of this section, the latter can be explained as follows. Although a good setting for the mutation rate does help the algorithm as a whole, on an individual basis the mutation rate does not directly influence a solution such that those with an 'optimal' setting have necessarily a higher fitness. Therefore, what will happen is that the mutation rate converges to the average of the initialisation range, in our case 2.5%. This was observed to be happening in the experiments. However, for our particular problem a mutation rate of 1.5% yields better results than a rate of 2.5%. Thus, the overall worse results with adaptive mutation are due to the mutation probability being too high.

To further investigate what might have caused the better performance of the adaptive crossover a single run of three data files is shown. The three files are chosen one each from set 4 (all constraints slack), from set 5 (some constraints slack) and from set 7 (all constraints tight). In Figure 3, the average probability of a crossover type against the current generation is plotted. This reveals some interesting facts and indicates why the strategy works. Towards the end of the search for all three files, C1 crossover is preferred. Presumably, this is because the search is close to converging and a less disruptive crossover, which leaves big chunks of individuals intact, is performing best.

Furthermore, there is a clear tendency in the graphs that the more difficult the file is to optimise, in our case because of tighter constraints, the longer the more disruptive and more flexible PMX and PUX operators were used. Particularly for the file of set 7, the C1 operator hardly features at all during the first half of the search because its use would be too 'conservative'. The graphs also show that generally PUX is preferred over PMX and that 'harder' problems require more generations to be solved.

## 5. Conclusions

As for the nurse scheduling problem reported in Aickelin and Dowsland (2000), the indirect approach proved to be superior to the direct genetic algorithm optimisation. However, the differences were not as great due to the way the data was created with an emphasis on less tight constraints making it easier for the direct approach to reach feasibility. Nevertheless, one might argue that the indirect algorithm and in particular the decoder was supplied with far more problem specific information than the direct genetic algorithm. And indeed, this is true. However, this was done because the indirect approach lends itself more easily to the inclusion of such information than the direct algorithms. This is an important lesson to be learnt from this research, although it would be premature to conclude that the indirect approach is always superior without further research.

With the indirect genetic algorithm and decoder, it was relatively easy to incorporate problem specific information, particularly in the form of making active use of constraints rather than as 'dumb' background penalties only. This again showed the advantages of this type of optimisation in comparison to the traditional genetic algorithms. The only problem faced was due to the sheer number of problem specific features that were included in the decoder. Therefore, suitable weights needed to be found to achieve a balance between exploitation and feasibility.

This was achieved by having some extra genes representing the weights and leaving the choice to the genetic algorithm via some special operators. Allowing the genetic algorithm to also choose the type of crossover employed extended this idea. The results found were of excellent quality with 100% feasibility and further improved rent, which was within 15% of a very optimistically set upper bound. Another advantage is that these results were achieved without the need for lengthy and possibly complicated parameter tests to find 'ideal' penalty weights and crossover rates. Moreover, these dynamic parameters have the benefit to be able to adjust with the search as has been demonstrated for the crossover rates.

To prove the general suitability of these ideas, they were re-applied to the indirect nurse scheduling approach. Again, the results (not reported here, but details can be found in Aickelin (1999)) were of better quality than before without any parameter tests being required. This indicates that there is potential in using this type of score assigning decoder plus a permutation based genetic algorithm for other scheduling and similar problems. The principles used in this decoder are widely applicable: Schedule one candidate at a time, measure the remaining slackness of constraints, identify positive and negative contributions to the objective and finally assign a weighted score. Of course, finding good weights will be difficult, but as shown in this research it can be left to the genetic algorithm to sort the weights out. Clearly, the direct-indirect question has not been resolved yet and this is a very promising area for further work. Already research is under way to apply the indirect genetic algorithm to set covering and partitioning problems with promising results in comparison to previous direct approaches.

| Parameter / Strategy | Direct Approach | Indirect Approach |
|---|---|---|
| Population Size | **1000** | **100** |
| Population Type | Generational | |
| Initialisation | Random | |
| Selection | Rank Based | |
| Crossover | Uniform with p = 0.66 | PUX with p = 0.66 |
| Mutation Probability | 1.5% per bit | 1.5% per bit for a swap |
| Replacement Strategy | Keep 10% Best | |
| Stopping Criteria | No improvement for 30 generations | |
| Penalty Weight | 20 | |
| Average Run Time per Instance | 19.9 seconds | 22.8 seconds |

Table 1: Parameters and strategies of the direct and indirect genetic algorithms.

| Weight | Low | Medium | High |
|---|---|---|---|
| $W_1$ | 500 | 500 | 500 |
| $W_2$ | 1000 | 1000 | 1000 |
| $W_3$ | 100 | 250 | 1000 |
| $W_4$ | 200 | 500 | 2000 |
| $W_5$ | 200 | 200 | 200 |
| $W_6$ | 2000 | 2000 | 2000 |

Table 2: Three types of weight settings for the mall problem decoder.

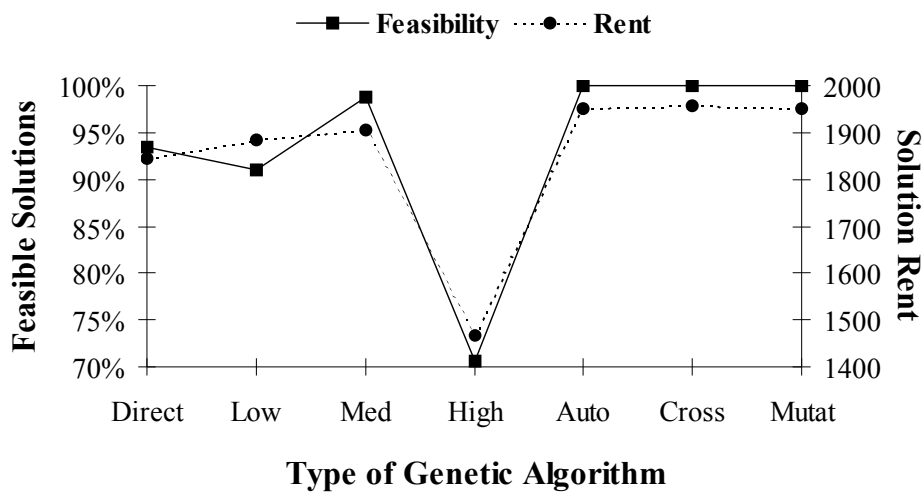

Figure 1: Comparison of results of various genetic algorithm approaches to the mall problem.

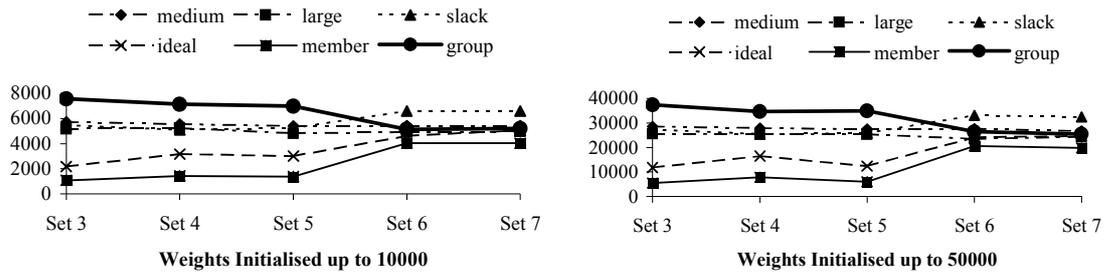

Figure 2: Different initialisation ranges of the decoder weights for the automated weight approach.

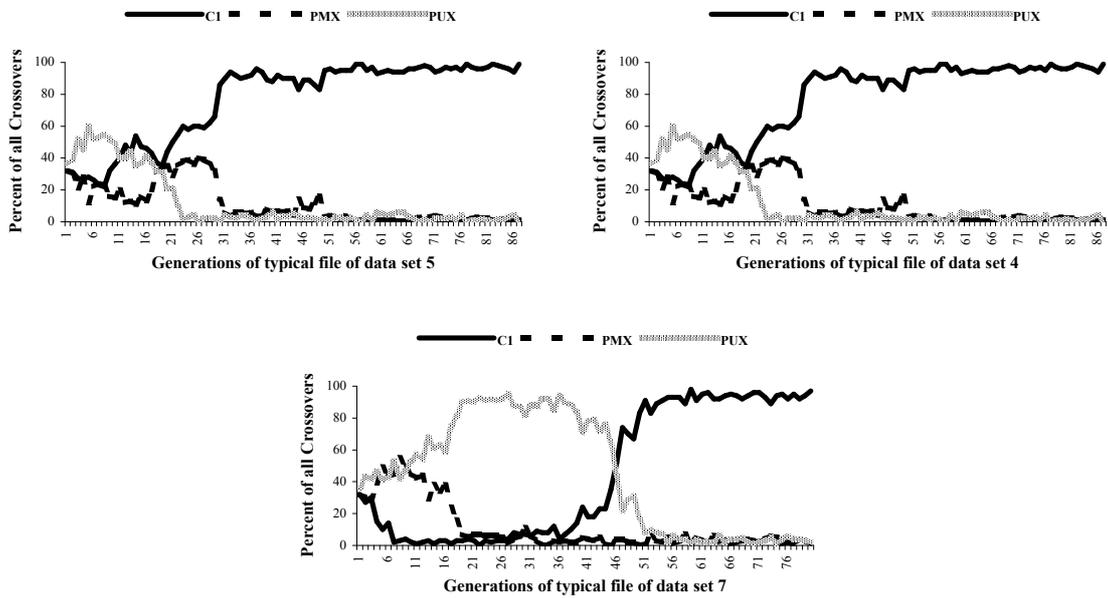

Figure 3: Comparison of crossover rates for three data files of increasing difficulty.